\documentclass[runningheads, twocolumn]{llncs}
\usepackage[textwidth=18cm]{geometry}
\usepackage[T1]{fontenc}
\usepackage[utf8]{inputenc}
\usepackage{lmodern}
\usepackage{graphicx} 
\usepackage[hidelinks]{hyperref} 
\usepackage[nottoc]{tocbibind}
\usepackage{authblk}

\newenvironment{packed_enum}{
\begin{enumerate}
  \setlength{\itemsep}{0pt}
  \setlength{\parskip}{0pt}
  \setlength{\parsep}{0pt}
}{\end{enumerate}}

\usepackage[english]{babel}
\usepackage{multicol}

\title{Word-wise intonation model for cross-language TTS systems}
\author{Tomilov A.A.\inst{1} \and
Gromova A.Y.\inst{1} \and
Svischev A.N.\inst{1, 2}}
\authorrunning{A. Tomilov et al.}
\institute{STC, Ltd, St. Petersburg, Russia;
\email{tomilov@speechpro.com} \and
ITMO University, St. Petersburg, Russia
 }

\providecommand{\keywords}[1]
{
  \small	
  \textbf{\textit{\textbf{Keywords}:}} #1
}

\begin{document}
\maketitle

\begin{abstract}
In this paper we propose a word-wise intonation model for Russian language and show how it can be generalized for other languages. The proposed model is suitable for automatic data markup and its extended application to text-to-speech systems. It can also be implemented for an intonation contour modeling by using rule-based algorithms or by predicting contours with language models. The key idea is a partial elimination of the variability connected with different placements of a stressed syllable in a word. It is achieved with simultaneous applying of pitch simplification with a dynamic time warping clustering. The proposed model could be used as a tool for intonation research or as a backbone for prosody description in text-to-speech systems. As the advantage of the model, we show its relations with the existing intonation systems as well as the possibility of using language models for prosody prediction. Finally, we demonstrate some practical evidence of the system robustness to parameter variations. 

\end{abstract}


\section{Introduction}

Nowadays TTS(Text-to-Speech) systems have largely achieved human-level expressiveness and naturalness \cite{budzianowski2024pheme}, \cite{ju2024naturalspeech}. One of the key parts of modern TTS is a prosodic model \cite{li2023styletts}. Prosody refers to a number of features: pitch, rhythm, pausation and stress. Additionally, prosody includes a part of paralinguistic features of a language which allow for expression of mood, state and intention. 

In the present paper we are focusing on a melodic component of prosody for a cross-language case (when one speaker should speak expressively in any of the supported languages)\cite{budzianowski2024pheme}.

There are approaches that are based on the output or hidden layers of language models based on BERT \cite{ju2024naturalspeech}, \cite{chen2021speech}, \cite{Liu_SSL_prosody}. Significant success in prosody plausibility is often reached at the expense of controllability. On average, prosody is expected to be successfully implemented. And if that is not the case, instruments for its adjustment are often somewhat implicit. One of the options for the ability to control emphasis is based on extracting pitch from audio and on unsupervised learning \cite{zhong2023eetts}. Another approach is handling synthesis parameters with text descriptions \cite{guo2022prompttts}, \cite{guo2023prompttts2}. Also in works \cite{Latif_e2e_prosody}, \cite{stephenson22_interspeech} it was shown that a nucleus shift often requires explicit markup. 

The main challenge is to develop an approach that provides  simultaneously controllability and improvement in expressiveness. For such task we propose a method based on universal description of pitch movements for every word in a phrase. We call it an Intonation PAttern-STAte (PASTA) model. Further we will show that with this method we can get an automatic, interpreted and rather simple description of melody relying only on word boundaries. This method enables manual control with the help of INTSINT (INternational Transcription System for INTonation) \cite{intsint_1} to which we can map other intonation systems (ToBI \cite{intsint_tobi}, ToRI or their variations, for example \cite{TL_ToBI}). Also we assume that it is possible to predict intonation model markups (in terms of a PASTA model) for a word or a token by using language models like BERT. It is harder to do that for the INTSINT markup due to symbols' irregularity over word positions (a single word can have few marks or not have them at all).

\textit{Our contribution is as follows}:
\begin{packed_enum}

\item we have presented an automatic controllable melody markup system;
\item we have demonstrated options for synthesis control with the proposed model (rule-based as well as based on BERT language models);
\item we have analysed the model's applicability to cross-language TTS;
\item we have collected and analyzed pitch patterns in Russian to show properties of the proposed model.  

\end{packed_enum}

\section{Model's description}

Our model implements a Momel (\textbf{Mo}delling \textbf{mel}ody) algorithm \cite{Hirst_momel} developed for analysis and synthesis of intonation patterns along with INTSINT. It helps to simplify pitch curve by using special spline approximation. 

By applying normalization one can achieve standardized patterns and an averaged level of a normalized pitch related to a phrase. By performing next stage, levels (rows in Fig. \ref{state_move}) and pitch patterns (at the bottom in Fig. \ref{state_move}) clustering it is possible to obtain quantized labels instead of a continuous set of lines.

We named this pitch pattern processing model an Intonation PAttern-STAte (PASTA) model. Here the state would mean an average pitch pattern position relative to the mean pitch level of a phrase. The pattern is the cluster which is the closest to the given normalized pitch pattern.

The next subsections describe each step in detail.

\subsection{Momel splines}
The Momel algorithm is based on the analysis of macromelodic and micromelodic components of speech. The former are modeled with a quadratic spline function (Fig. \ref{dtw_ex}). A derivative of an approximating function equals zero at the spline knots. The spline spreads on unvoiced parts of a contour excluding pauses. The algorithm defines significant points of pitch changes called Momel anchors. The advantage of this approach is that the defined points are relatively speaker-independent, content-independent and language-independent. Thus in theory it is possible to analyze pitch regardless of the language \cite{hirst1998intonation}.

\begin{figure}
	\includegraphics[scale=0.3]{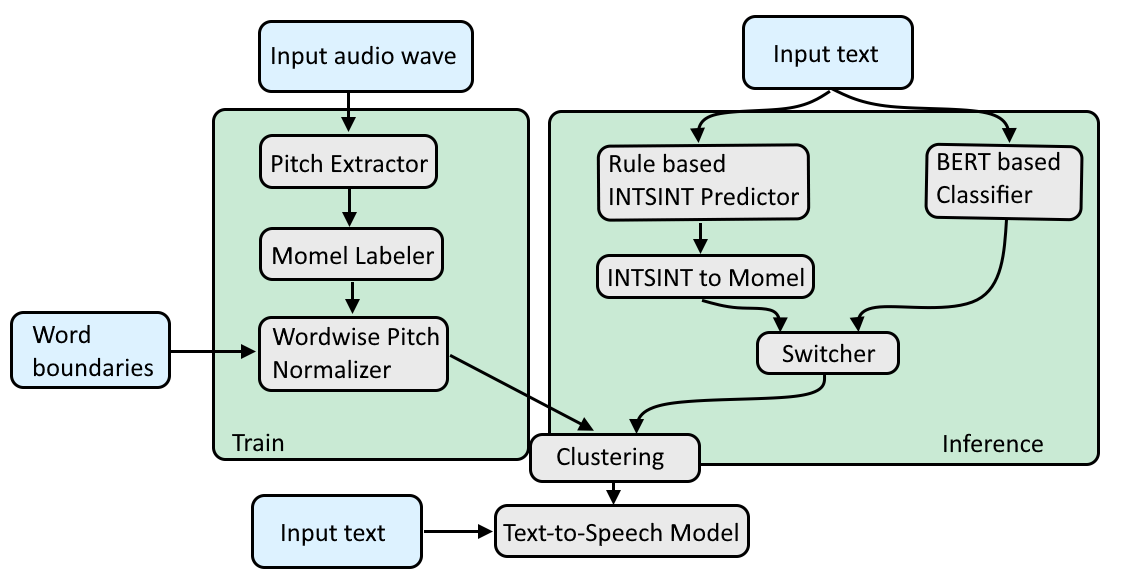} 
	
	\caption{Overall data processing scheme}
        \label{whole_scheme}

\end{figure}

Intonation analysis is often considered by dividing phrases into intonation units (IUs). These consist of one or more words and are generally connected with semantic division or intonation organization of an utterance \cite{Roll_2023}. It is a hard task to mark contour boundaries and distinguish their types automatically based on an audio. It is especially complicated to mark a nucleus of an IU which is also called an intonation centre. For emphatic constructions it may be relatively easy to locate nucleus position \cite{peng2024empirical}. An analysis of IUs relies on a manual markup which is often ambiguous and relies sometimes on a speaker's intention. Therefore, it requires complex analysis of both speech and text \cite{Rosenberg2010AutoBI}.

\subsection{Obtaining $F_0$ patterns}

\begin{figure}
	\includegraphics[scale=0.35]{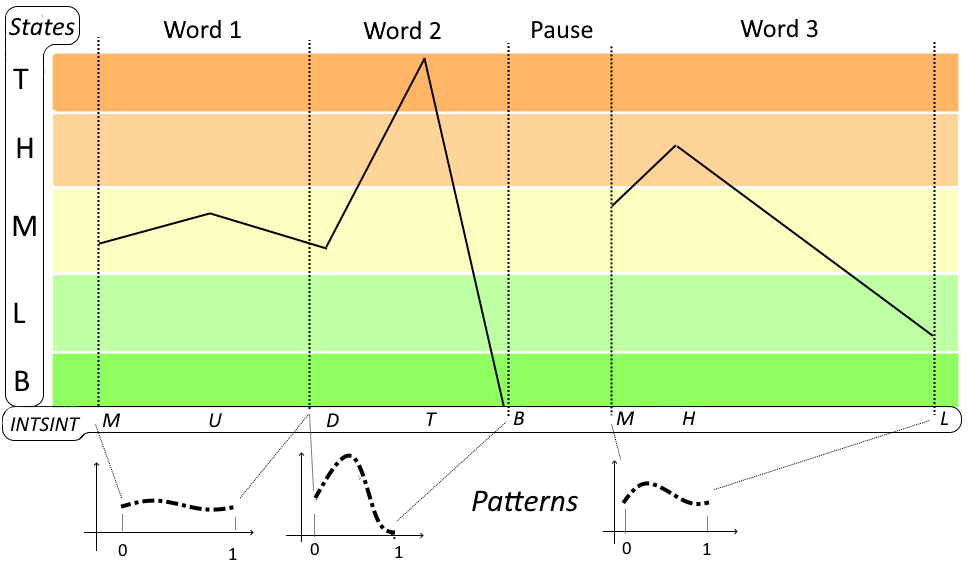} 
	
	\caption{The scheme of an Intonation \textbf{PA}ttern-\textbf{STA}te model. The state is an average pitch pattern position relative to the mean pitch level of a phrase. The pattern is the label of the closest cluster.}
        \label{state_move}

\end{figure}

As mentioned previously, it is difficult to determine boundaries of IUs. Therefore we propose to analyze intonation in terms of words. By considering both audio and text we can get word boundaries with Montreal Forced Aligner (MFA) \footnote{https://pypi.org/project/Montreal-Forced-Aligner/} or WhisperX \cite{bain2023whisperx} \footnote{https://github.com/m-bain/whisperX}. On the basis of automatically obtained $F_0$ curve (Crepe \footnote{https://pypi.org/project/crepe/} or RMVPE \cite{Wei2023}) it is possible to get marcomelodic component on a word by applying the Momel algorithm. Then by normalizing time and $F_0$ (Fig. \ref{state_move}) one can receive normalized $F_0$ patterns. By time normalization we mean $F_0$ approximation by a line with a fixed number of points (usually evenly spaced, i.e. in compressed or stretched time). $F_0$ is normalized with dividing $F_0(t)$ contour by a $\langle F_0 \rangle$ constant (eq. \ref{f0_norm}), which is an average pitch value for either this speaker (on the entire corpus) or this particular phrase. Thus we reduce each pitch movement to its "canonical" form in terms of a model.

\begin{equation}
F_0(t) = \frac{F_0(t)}{\langle F_0 \rangle}
\label{f0_norm}
\end{equation}

It should be mentioned that according to phonetic studies pitch movements are often starting or are defined on a stressed vowel \cite{hirst1998intonation}. It is particularly characteristic of a nucleus. What is important for our further discussion is the fact that similar patterns will differ in \emph{a place of extremum or turning-point}; otherwise, they will resemble each other.

The normalization allows to form word pitch patterns in a corpus as a number of vectors: $[N \times N_{F_0}]$, where N is a number of words in a corpus, $N_{F_0}$ is a number of points selected for time normalization (Fig. \ref{state_move}). Afterwards we can perform clustering, for example, with KMeans. Various functionals, for instance, Euclidean distance can serve as a KMeans metric (a measure of a cluster proximity). As intonation can vary because of stress \cite{stress_nucleus_1}, it makes sense to consider sustainable clustering metrics, for example, based on a dynamic time warping (DTW) or SoftDTW \footnote{https://pypi.org/project/tslearn/}. Algorithms based on DTW take into account that a curve normalized along the y-axis can be deployed along the x-axis (Fig. \ref{state_move}) so two pitch curves can be associated with a single cluster center despite the possible shifting of a main cluster pattern along normalized time. The main hypothesis is that pitch movements (particularly in a word carrying a nucleus) act in a similar way on a stressed syllable, before it and after it. On the other hand, the movement after the stressed syllable may be absent in a word with the stress located on the last syllable. This way it is possible to model a pitch movement that is structurally different but semantically connected with a movement which is more typical for these prosodic conditions.

\subsection{The choice of a word-wise model}

Phonological models of Russian have traditionally considered intonation in terms of a limited number of certain intonation patterns. These patterns spread over an intonation unit associated with a pitch contour. These models view prosodic word only as a component part of an intonation unit that carries a word stress. Phonetic intonation models can reflect almost all pitch movements comprising intonation contours. At the same time phonological models focus mainly on pitch movements carrying the most prominence in an intonation unit, i.e. its nucleus.

Research shows that in Russian on average 3 out of 10 words are prosodically prominent \cite{Hart_Collier_Cohen} (but in certain regional variants of Russian pitch accents may be more frequent ("word-by-word melodic contour") \cite{knyazev_evstigneeva}). That is why in Russian (and in English), if pitch clusters on a word are modeled, they will reflect prosody of words that are marked by pitch as well as of words that are unmarked by it. Therefore, a high percent of clusters should potentially have a pitch curve of a level tone or a slight rise/fall (it may be seen from a number of samples of 'slow' clusters in Fig. \ref{app_clusters}).

Lobanov's work \cite{lobanov_1} explores intonation patterns and shows that it is possible to define dozens of intonation patterns that retain a rather similar contour after a change of a speaker or a language. Unlike in our model, these intonation patterns can stretch over one word as well as a number of words. This makes an automatic markup rather complicated.

Therefore, our model can be viewed as a number of components, in a way a building material for intonation patterns described in other intonation systems.

\section{Pitch patterns clustering}

In this section a clustering model of pitch patterns is described which can be used for intonation markup (Train part in Fig. \ref{whole_scheme}). Also we consider the model's application for English, Russian and Kazakh.

\subsection{Pitch processing}

In order to simplify pitch modelling we have made some assumptions:
\begin{packed_enum}
    \item pitch curve can be decomposed into macromelodic and micromelodic components with the former related to intonation patterns;
    \item we can artificially continue pitch curve on unvoiced parts.
    
\end{packed_enum}

These assumptions lead to a Momel system which models pitch within a phrase with a number of quadratic splines. The real $F_0$ curve is approximated using these splines. To describe the resulting spline, it is enough to know the location of knots (Momel anchors) as a pair (time, frequency) or ($t_i, F_{oi}$). Outside of the domain in order to avoid nonphysical values we suppose that the spline has a constant value.

As a spline is composed of quadratic functions, any time slice of this spline (corresponding to a word position) will also be a set of quadratic splines. It is worth mentioning that after spline approximation any manipulations with time normalization are done trivially without losing precision.

If melodic features within a word change relatively slowly, it is possible to upsample or downsample such an array that describes this pattern without the loss of information. If $F_0$ features have rather complex structure, a transfer to a sparse grid leads to losing information, according to the Nyquist–Shannon theorem. Then, in order to keep the main curve we can perform downsampling with averaging after applying low-pass filter to the signal \footnote{https://flylib.com/books/en/2.729.1/decimation.html}.

\subsection{Euclidean metric's limitations}
If we cluster slices of Momel splines (corresponding to words) with a Euclidean metric, there will be many various curves depending on a number of selected clusters. The most common ones will be lines with a different incline. 

It is worth mentioning that pitch curves may have shifts in extremums and steepness positions that may not be related to significant prosodic difference. The reason is that the main movement (extremum, turning-point, maximum line steepness) is often characteristic of a stressed syllable in a nucleus. And many languages have lexical stress, i.e. word stress does not fall on any particular syllable but rather is lexically encoded. A shift of a word stress often does not change an intonation pattern in terms of representing a certain communicative type \cite{lobanov_1}. Then a Euclidean metric is suboptimal. On the other hand, using DTW \footnote{https://pypi.org/project/tslearn/} can help to ignore feature position shifting relatively to a word centre. Thereafter, we can still 
form a cluster's barycentre to assess its dynamics (for more examples see \ref{app_clusters}).
This is why the Euclidean metric is more suitable for state clustering rather than for move clustering. 

\subsection{DTW clustering}

\begin{figure}
\begin{center}
	\includegraphics[scale=0.55]{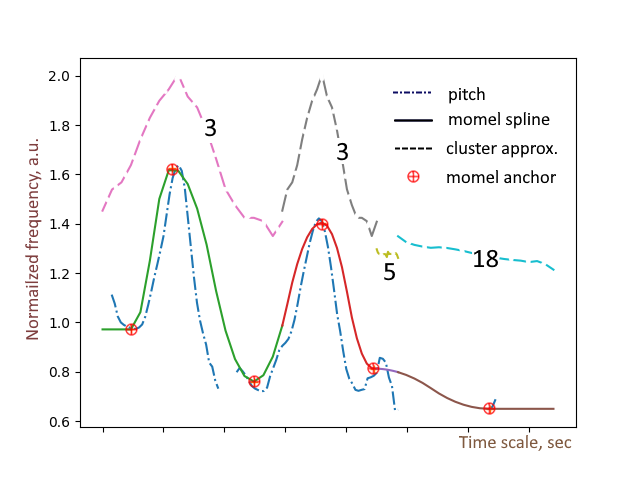} 

	\caption{An example of $F_0$ approximation with a Momel spline and a DTW clustering. Clusters on a word can be stretched or compressed by time. Numbers indicate a cluster number.}
 	\label{dtw_ex}
	\end{center}

\end{figure}

When performing a clustering, a unit of processing is a slice of a Momel spline that corresponds to a particular word in the pronunciation time. After a Momel spline slicing from the word start to the word end we can expect that such cut will be also approximated by a combination of quadratic functions. It may be shown using the definition of a Momel spline and a Heaviside function as a slicing operator:

\begin{equation}
F_{momel}(t) = \sum_{n=1}^{N}{Q_{spline}(t_n, F_{0n}) \times (H(t_{st}) - H(t_{end}))}
\label{momel_eq}
\end{equation}

where  $Q_{spline}(t_n, F_{0n})$ is a pair of quadratic functions defined between Momel anchors (according to \cite{Hirst_momel}), $H(t)$ is a Heaviside step function.

\begin{figure}

    \includegraphics[width=.35\textwidth]{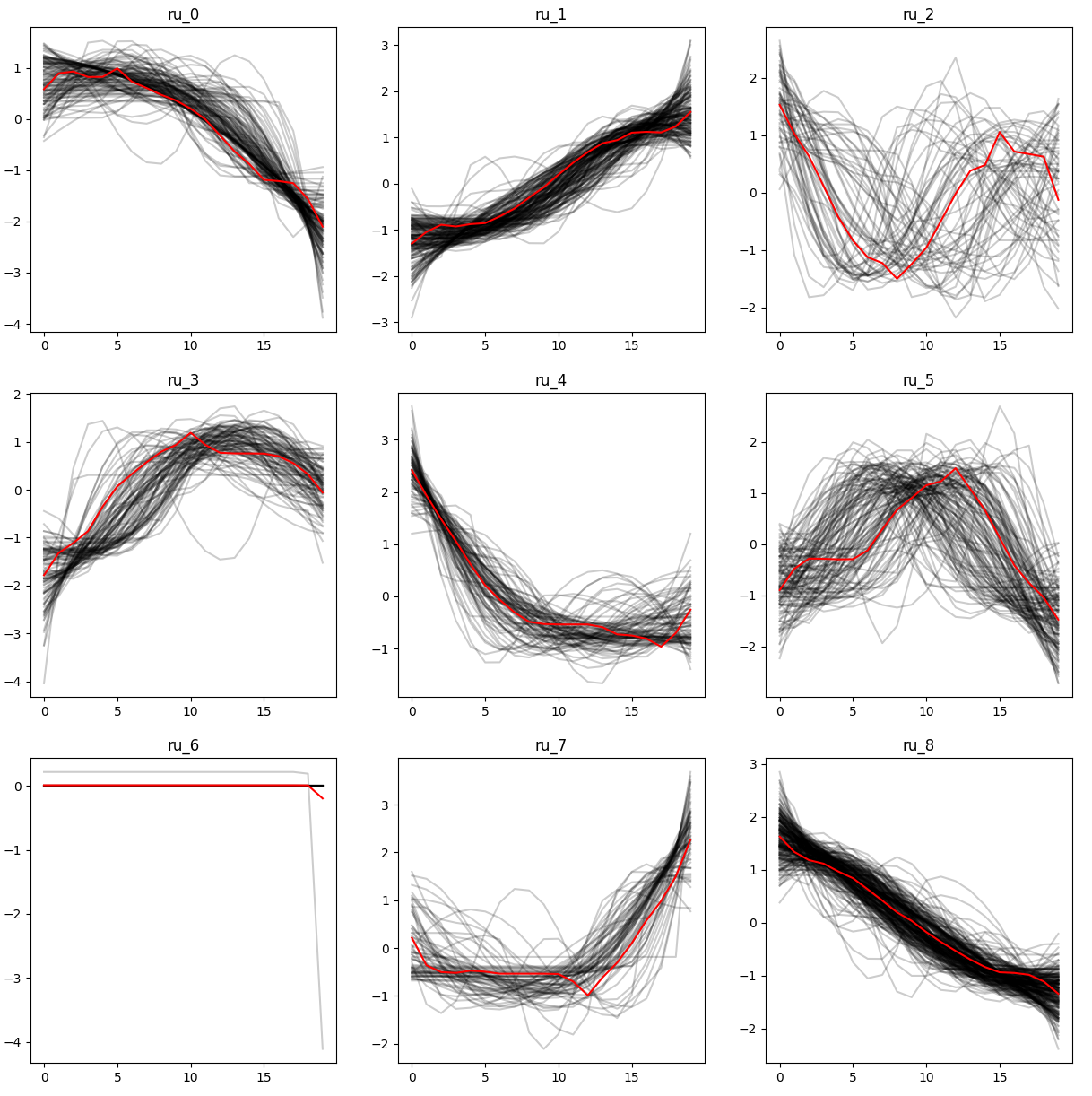} \vfill
    \includegraphics[width=.35\textwidth]{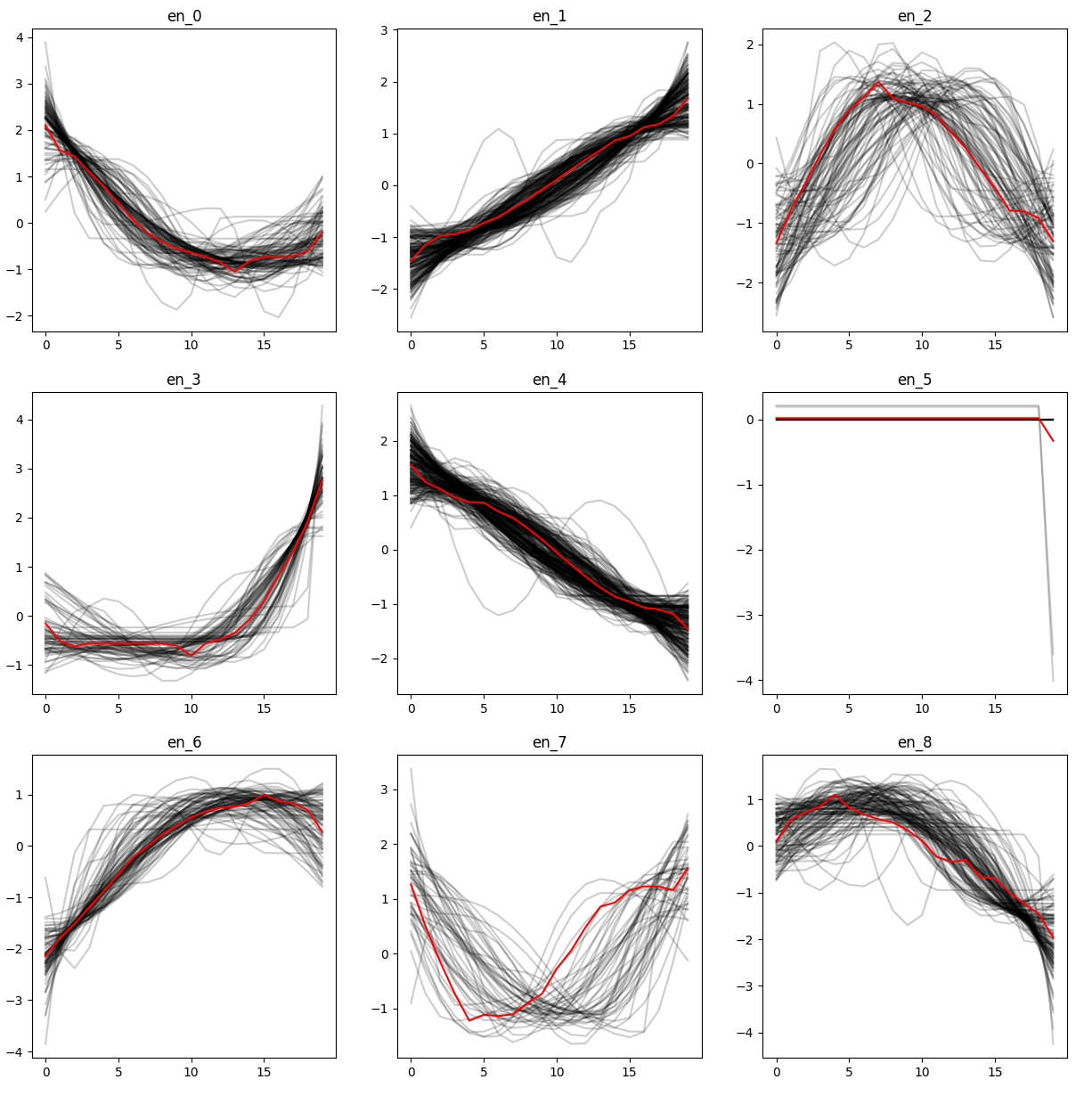} \vfill
    \includegraphics[width=.35\textwidth]{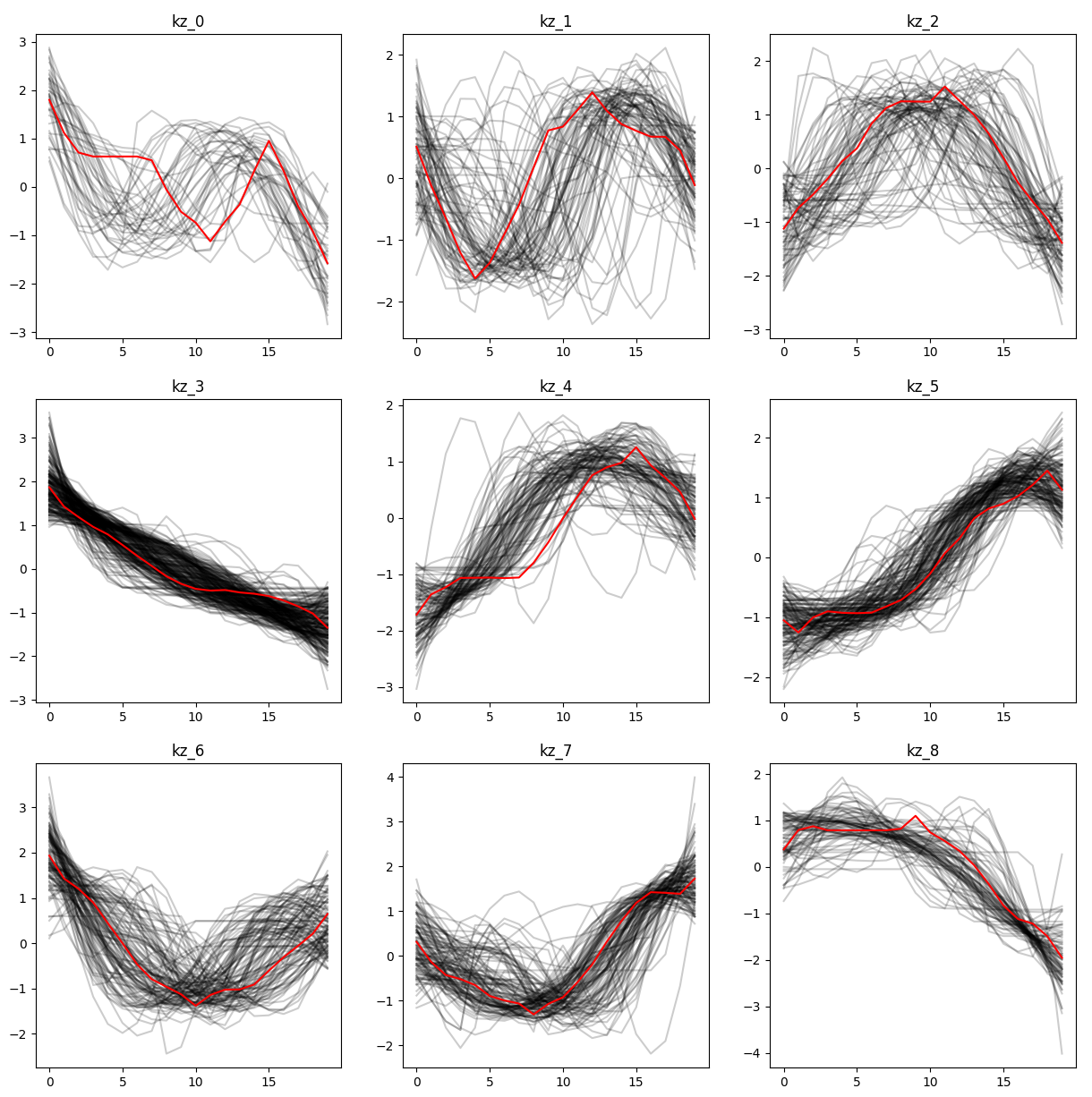}
    \caption{An example of a word-wise clustering for Russian, English and Kazakh (from top to bottom). Instances for each cluster are indicated by black. Clusters' barycentres are indicated by red.}
    \label{dtw_ru_9}

\end{figure}

By clustering into a different number of parts it is possible to obtain certain patterns. Having analyzed them we are able to distinguish the ones that are the most frequent and at the same time occurring independently for various languages. Above all, these include linear movements (for example, clusters '1' and '8' in Fig. \ref{dtw_ru_9}, left plot). Then there are 4 types of half parabolas ('4' and '7' in Fig. \ref{dtw_ru_9}, left plot) and 2 types of parabolas ('5' in Fig. \ref{dtw_ru_9}, left plot, and '7' in Fig. \ref{dtw_ru_9}, right plot). 

It enables to suggest that:

\begin{packed_enum}
  
    \item Russian, English and Kazakh mostly have the same intonation patterns within a word;
    \item with subsequent consideration of other languages the main patterns will remain the same but the new ones might be added.
\end{packed_enum}

Fig. \ref{dtw_ex} demonstrates, that the proposed system on the base of a DTW clustering has enough expressiveness to resemble the real $F_0$ curve. After all, in the end a basically simple one curve is obtained (a sequence '$3\rightarrow3\rightarrow5\rightarrow18$' on the Fig. \ref{dtw_ex}).

\section{Intonation synthesis}

An intonation synthesis is a procedure of a conversion from a text to an intonation markup representation with the help of either a language model or proxy systems like the INTSINT (the Inference part in Fig. \ref{whole_scheme}).

\subsection{Connection with intonation models}

One of the intonation's main functions is to convey a communicative type of an utterance. That is why for modeling an intonation contour it is important to retain intonation patterns responsible for differentiating between communicative types.

In a ToRI intonation system (a Russian adaptation of a widely used ToBI system) \cite{ode2008transcription} $C. Od\acute{e}$ describes 6 pitch accents: H*L, H*H, H*M, L*, HL* and L*H. They are used for defining prosody at the nucleus and at the adjacent syllables. The described pitch changes are perceptually relevant for this intonation contour. The ToRI system also presents a way of denoting a tone level at the intonation unit boundaries.

If we know the boundaries of an intonation unit and the nucleus location within those intonation units, we can directly transfer from ToRI to INTSINT \cite{intsint_1}. For example, as in ToRI, an initial tone of an intonation unit can be marked in INTSINT (T, M, B). INTSINT can also reflect pitch at the nucleus and at its adjacent syllables. There was an evidence for a high degree of accordance between ToBI and INTSINT in the type of a tonal movement \cite{intsint_tobi}. As for words without such prominence (features that are not so perceptually important for an utterance), marks B, S, M and T marks indicated a level of tone on these words. As in ToRI system, there is not always a need to mark the final pitch, if it is evident from the previous part of a contour. For instance, if a nucleus is at the final part of a contour, and the final pitch can be directly concluded from the nucleus mark.

For intonation of English a description in terms of a pre-head, a head, a nucleus and a tail has been historically used \cite{o'connor1973}. This system requires not only 
intonation unit's boundaries and a nucleus location but also we need to know if the word is accented or not. This may be helpful in defining where a pre-head ends and a head starts (if there are these parts in this intonation unit). 

Having all previously mentioned information, we can model all parts of an intonation contour. A time value is appointed to every given mark. The time value depends on a location of a sound that is intended to have this certain pitch. If the target mark denotes not a pitch move but rather a pitch level tone, and only a tone level is important for this contour, we put this mark on a time value corresponding to a stressed vowel of the word. For example, in Russian words after a nucleus have a low pitch in most cases.

Later on, in order to transfer to an Intonation PAttern-STAte system, according to recursive rules \cite{intsint_1}, it is necessary to put corresponding INTSINT marks to the set position of pseudo-time. In order to get pseudo-time an a priori assumption for phonemes' length should be introduced. This can be done by calculating a mean length of a phoneme, and then set them for inference. But we will try to incorporate phoneme durations as well as clusters to TTS model during end-to-end train later.

A key (a mean value) can be equaled to 1 ($key=1.0$) because of the normalization of initial data according to Eq. \ref{f0_norm}. Supposing that pitch has a variation in 1 octave (so that $\frac{min(F_0)}{max(F_0)} = 0.5$), after the frequency normalization (eq. \ref{f0_norm}) the absolute variation is $range=\frac{2}{3}$. Then a normalized Momel spline can be formed. Next, knowing time values of a word start and a word end, we can define pitch on every word according to Eq. \ref{momel_eq}. Afterwards a DTW clustering is performed and an intonation markup with cluster numbers is obtained.

\begin{figure}
\begin{center}
	\includegraphics[scale=0.5]{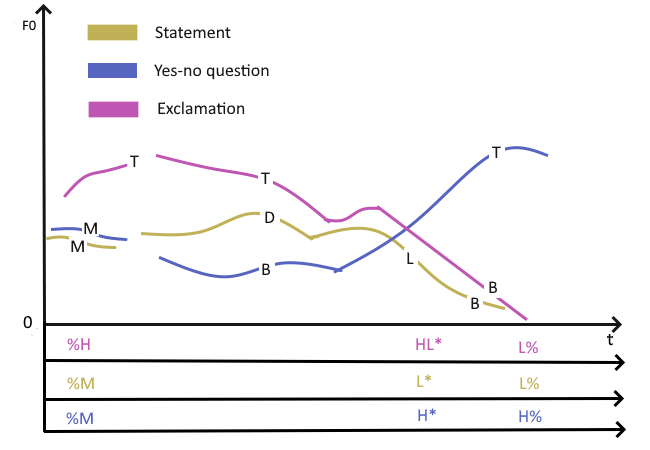} 
	\label{contours}
	\caption{An example of INTSINT and ToRI intonation markups for a statement, a question and an exclamation in Russian}
	\end{center}

\end{figure}

In Russian the most common contours for a statement, a yes-no question and an exclamation correspond to these pitch accents: L*, H*L and HL* respectively. By transforming these marks into INTSINT, we get L, TL (or HL and T or H if there are no post-nucleus syllables) and TL (or TB) with a corresponding stressed syllable location (Fig. \ref{contours}).

\subsection{Language model for intonation prediction}

As has been shown earlier, a sequence of word-wise pitch clusters reflects prosodic features of an utterance. On the other hand, the works \cite{chen2021speech}, \cite{c6906d38f66b4881848463e3eef3f3dc},  \cite{devlin2019BERT} demonstrate that BERT-based models adopt prosodic information during semi-supervised learning. Thus it may be assumed that with only a text information we can predict (with some quality different from random) cluster sequences within an analyzed model. These sequences can be used to improve prosody in TTS.

In this case we get an interpreted markup and an instrument for an explicit control through an INTSINT proxy-model that focuses on a prosodic aspect of speech. It is important to mention that a TTS model can learn in connection with a cluster system on the whole dataset. And a fine tuning of a prosodic cluster classifier can be performed on a wide spectre of pretrained BERT models and a pre-selected target data domain (for example, a dialogue or spontaneous speech). This provides a significant flexibility compared to a direct injection of certain BERT features into a model.

\subsubsection{Experimental settings}

We have chosen BERT-based pretrained models of different capacity: RuBert tiny \footnote{https://huggingface.co/cointegrated/rubert-tiny}, RuBert tiny 2 \footnote{https://huggingface.co/cointegrated/rubert-tiny2}, distill ruBERT small \footnote{https://huggingface.co/DeepPavlov/distilrubert-small-cased-conversational}. A small size is intended for a fast inference.

For datasets we used English and Russian parts of Common Voice \footnote{https://commonvoice.mozilla.org/ru}, Libri TTS \cite{libri_tts} and some proprietary datasets. Unlike in TTS, a minimal learning unit is a sentence. It is necessary for providing context. Data was filtered to exclude insertions in other languages.

It is worth noting that a cluster sequence is significantly speaker-dependent because of a different expressiveness and 
acceptable variations within an intonation pattern. That required a thorough data selection for learning.

Models' original tokenizers were used. A word's cluster mark was assigned to its every token.

To remove dependency on the punctuation, punctuation marks were not taken into account when summing up tokens.

We will assume that argmaX from a sum of predicted distributions of its token labels is a predicted cluster label. For a cluster system (Fig. \ref{app_clusters}) was chosen. BERT's weights were not frozen. A learning speed for a classifier is $lr=1e^{-3}$, for BERT it is $lr=1e^{-5}$.

\subsubsection{Results for a cluster learning for BERT models}

During tuning it has been found that it is rather difficult to reach a high accuracy for a cluster classifier learning. A detailed analysis shows that the errors generally were linked to predicting a cluster with the same direction but with different magnitude. For example, clusters 3 and 20 are close to a level tone. It is possibly associated with some redundancy. On the one hand, it provides an instrument for making a minimal adequate basis for a description of intonation patterns that are semantically different for a wide range of models (assuming that all of them are based on BERT but learned from different data and have different capacities). On the other hand, prediction errors usually do not affect strongly the correctness of an intonation pattern but rather its expressiveness.

Cases in which a confusion matrix shows significant difference between a predicted cluster and a target cluster and the fact that clustering is not robust to intonation at the boundaries leads to the conclusion that it is better to utilise a phoneme-wise or a syllable-wise model for an improved classification. This correlates to empirical evidence demonstrated by \cite{Phoneme_bert} on how BERT features affect prosody quality in TTS.

\subsection{Clusters' properties with parameter variation}

This section analyzes robustness of a certain cluster for a specific intonation pattern regardless of speaker's gender. We consider clusters' variability when changing a communicative type of an utterance and thus changing an intonation pattern.

\begin{figure}
\begin{center}
    \includegraphics[scale=0.3]{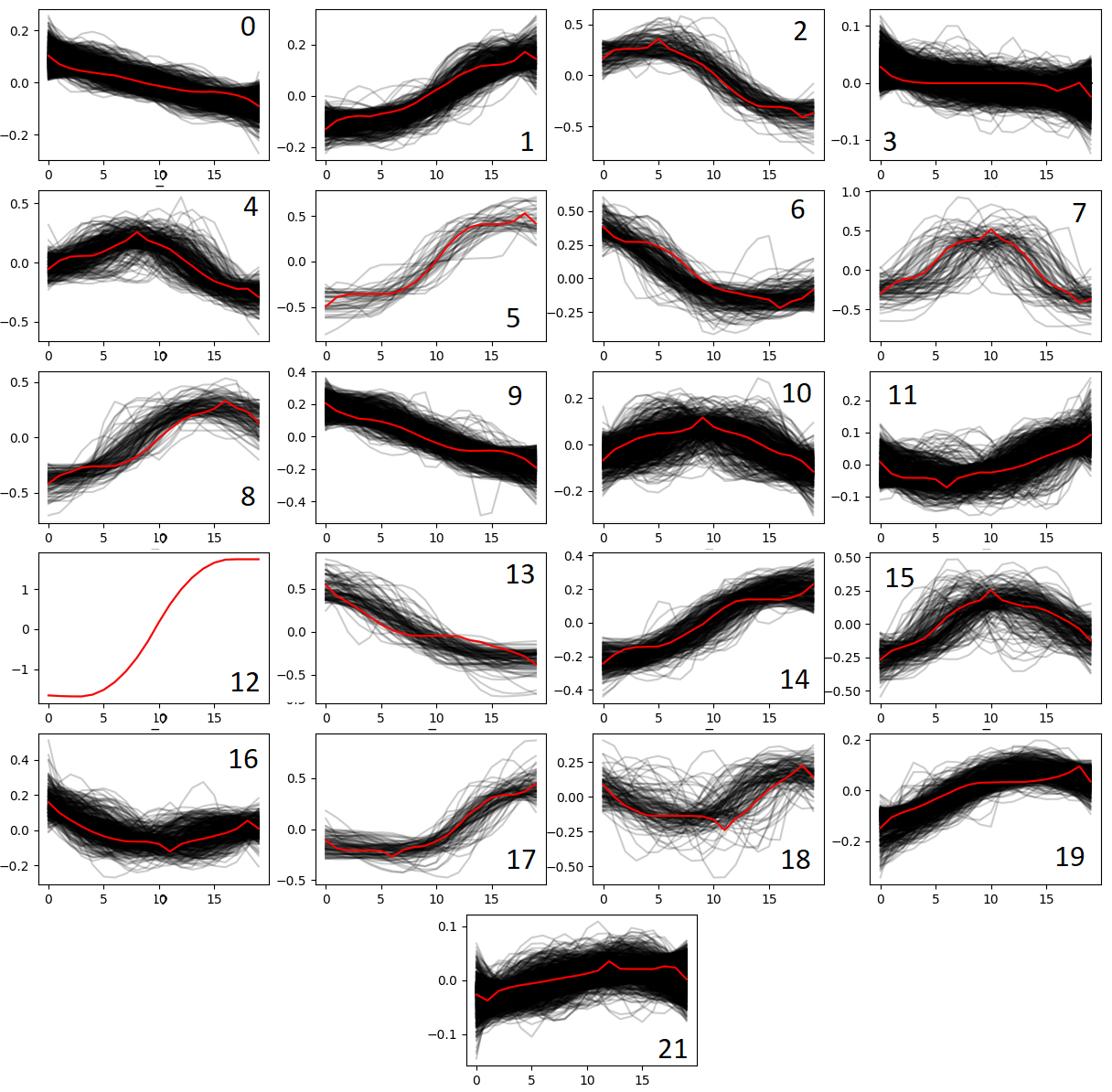} 
	
    \caption{Clusters' barycentres (in red) and movements belonging to a cluster from the database (in black). Patterns for clustering were collected by combining all three languages databases.}
    \label{app_clusters}
    \end{center}
\end{figure}

The initial hypothesis of implementing a DTW clustering was to provide cluster uniformity when shifting a stressed syllable within a word. We have conducted an experiment for Russian. The goal was to demonstrate contour variations on a word with a change of parameters. The said parameters were: 4 communicative types (statements, yes-no questions, continuatives and exclamations), a number of syllables in a word (from 1 to 5) and a position of a stressed syllable within a word. 8 (4 male and 4 female) Russian native speakers took part in an experiment. Speakers' age varied from 18 to 37. The speakers were asked to say one-word sentences (or parts of a sentence for continuatives). For each speaker the total number of words was 72 (18 for each intonation pattern). The punctuation marks corresponded to the target communicative type.
\begin{figure}
    \includegraphics[width=.37\textwidth]{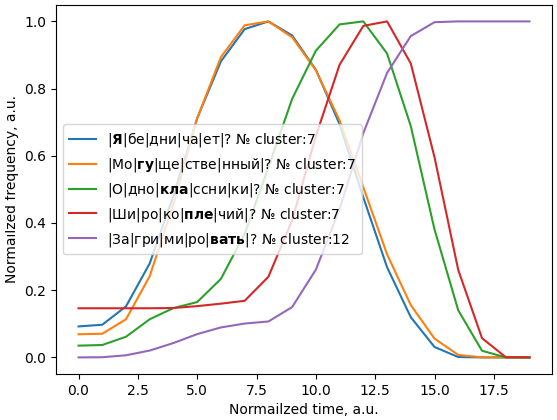} \hfill
    \includegraphics[width=.37\textwidth]{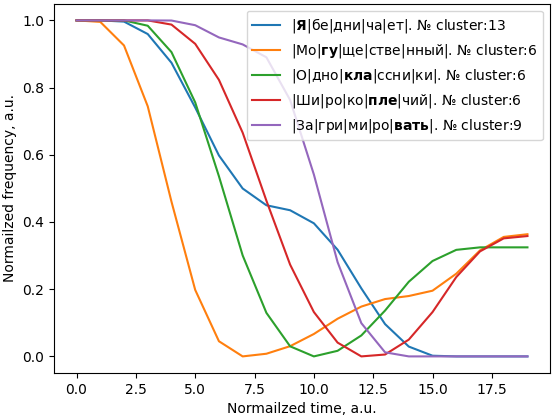} \hfill
    \includegraphics[width=.37\textwidth]{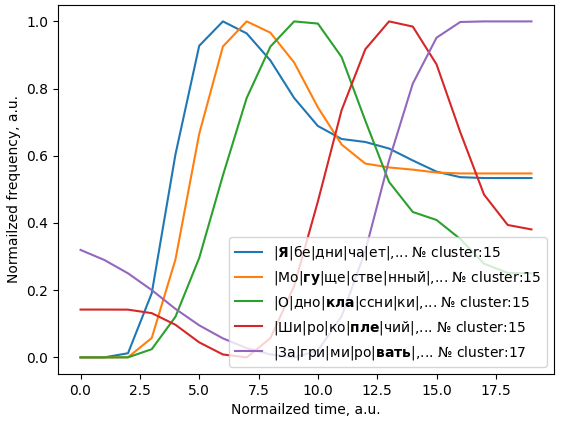}

    \caption{Common intonation contours for a 5-syllable \textbf{yes-no question} (top), \textbf{affirmative}(center) and \textbf{continuative}(bottom) utterances with a stressed syllable shift from the first to the last syllable}
 	\label{Stress_var_5_syll}

\end{figure}
All the audio samples were processed in accordance with the described normalization for frequency (eq. \ref{f0_norm}) and time (Fig. \ref{state_move}). After this procedure they have the same length and comparable magnitudes. Clustering among the chosen DTW clusters (Fig. \ref{app_clusters}) was also performed.

The study revealed a high degree of accordance for yes-no questions. The given example for 5-syllable words spoken by one speaker (Fig. \ref{Stress_var_5_syll}) clearly demonstrates a peak shift within a contour. This shift corresponds to a shift of a stressed syllable position. When the last syllable of a word is stressed, the contour is different. This is a common case for the statement yes-no question intonations (yes-no questions: Fig.\ref{Stress_var_5_syll} (left); statements: Fig. \ref{Stress_var_5_syll} (center) or continuation: Fig. \ref{Stress_var_5_syll} (right)). The reason for this is that there are no syllables left for a fall. If the word stress falls on a first syllable, we assume that the important part of an intonation contour can be located at the preceding syllable, i.e. at the preceding word. Therefore, there is a tendency of retaining a specific cluster for certain intonation patterns; however, there is a limitation for when the word stress is located on the first or on the last syllable.

\begin{figure}
    \includegraphics[width=.45\textwidth]{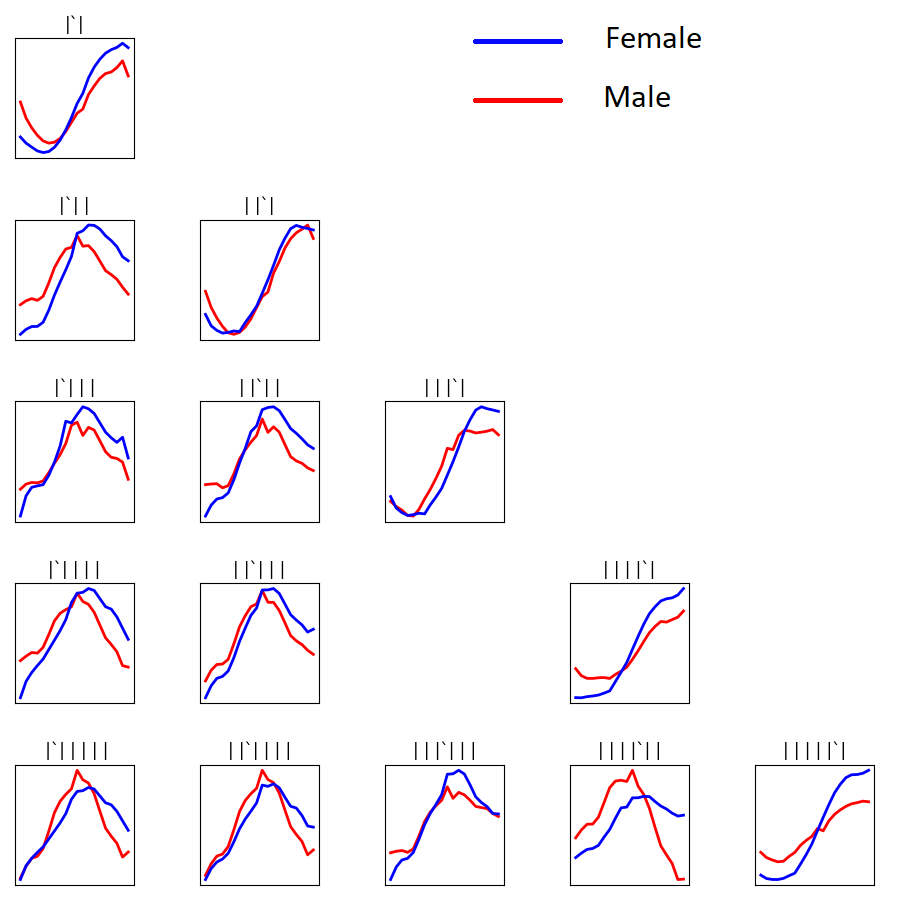} \hfill
    \includegraphics[width=.45\textwidth]{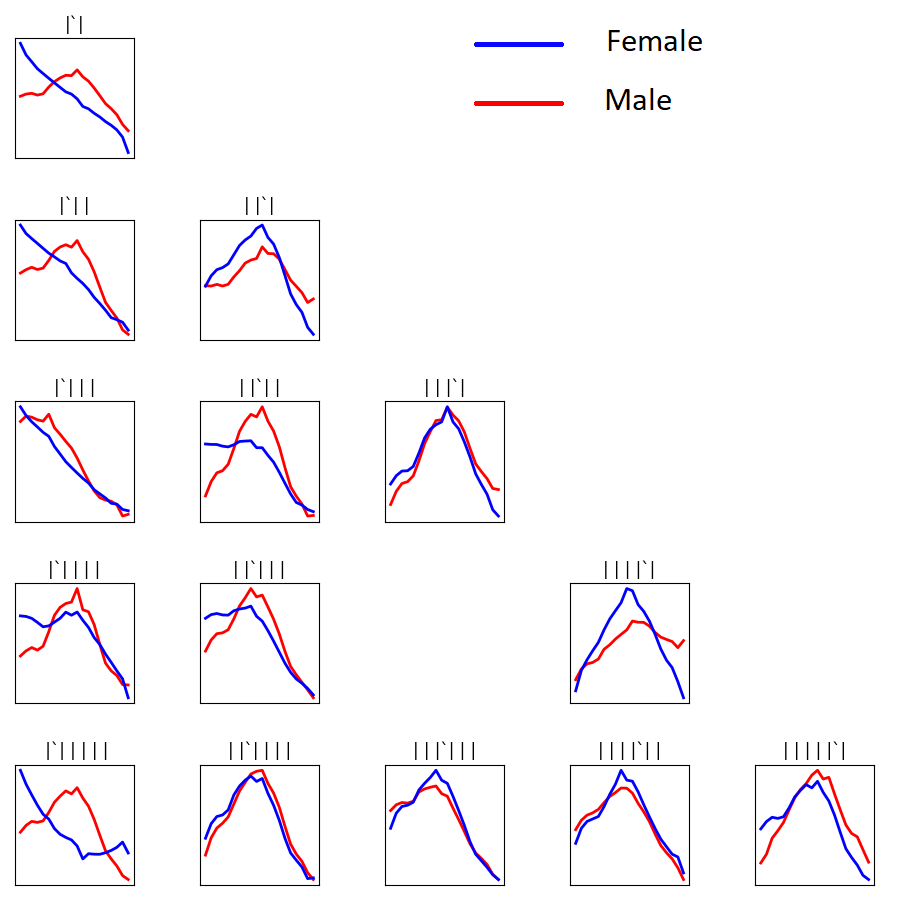}

    \caption{A variation of intonation contours for a one-word \textbf{yes-no question} (on top side) and \textbf{exclamation} (on bottom side) with a stressed syllable shift for a different number of syllables in a word for male (in red) and for female (in blue)} 
    \label{traingle_stress}

\end{figure}

To approve the hypothesis of the similarity of intonation patterns (with maybe an exception of the first and last stressed syllables in a word) we aggregate all variations over the corpus for the fixed intonation types (Figs. \ref{traingle_stress}). Number of syllables in a word variation depicted in row number from top to bottom. A stress syllable positions varies from left to right in columns. To analyze gender influence, we separately plot male and female speakers in red and blue correspondingly.

For a yes-no question (Fig. \ref{traingle_stress}) it is clearly seen that the position of a stressed syllable on the last syllable corresponds to a change of pattern as well as a number of cluster. An example of such a change may be seen in Fig. \ref{Stress_var_5_syll}. An independence of pattern changes over gender was revealed for the yes-no question intonation type.

Similarly, we can analyze exclamation intonation patterns (Fig. \ref{traingle_stress}). Contrarily to the previous type of intonation, the intonation patterns revealed less stability during parameter changes. As an example, males seem to have more stable intonation patterns over variations. On the other hand, females have a change in intonation pattern when the first syllable is stressed in a word. Such variations should probably be included into intonation models to be more realistic and plausible. 

\section{Conclusions and Discussions}

In the paper a word-wise intonation model is introduced that allows to represent a wide range of intonation patterns in a compact way. Thereafter, a prosody markup can be done with this model on the base of a pitch contour and word boundaries. Our research revealed that for all of the three examined languages there are similar patterns. Almost all pitch movements could be described in terms of that set of relatively common patterns. 

If there is some need to increase the capacity and expressiveness of the model it can easily be done by adding more clusters. Therefore, we have demonstrated the model's applicability to the cross-language case. The mechanism of adding new intonation patterns is rather transparent and can be implemented by just increasing a number of clusters. 

We have shown that using DTW as a distance measure in this case allows for more pitch variations' robustness in the model. It also decreases the influence of the stressed syllable's position on a cluster. It is especially noticeable in the case of the stress position close to the centre of the word. Moreover, it provides visualization of cluster's barycenters and therefore provides the whole model with interpretability. 

The advantage of the proposed model is that it can work for standard variants of the language as well as for spontaneous speech. The cluster prediction of the latter can be achieved only with using language models.

There is still much to be explored, including improving language model accuracy for the intonation cluster prediction. By the way, we have a hypothesis that incorporating information about a stressed syllable somehow (by using phoneme-wise or a syllable-wise BERT or by adding a syllable stress feature to a classifier) can significantly improve the cluster prediction accuracy.

It is worth noting that the proposed model does not take into account a stressed syllable's location relative to a cluster. That leads to ambiguities when one cluster may correspond to two semantically different intonation patterns. A TTS model can solve this ambiguity by adding locations of stressed vowels. However, without these locations the model would be incomplete. In our future works we will make efforts to solve this problem.

\section*{Acknowledgements}
We would like to express gratitude to our colleagues from the Speech Technology Center for their help in recording and processing a dataset for the experiment and for the useful discussions that significantly improved the present paper. The authors thank Sergey Novoselov, Artem Chirkovsky, Georgy Dobrikov, Eugeny Lukyanets, Marina Volkova, Ekaterina Anisina for their helpful discussions and feedback.

\bibliographystyle{ieeetr}

\bibliography{ssam}

\begin{thebibliography}{10}

\bibitem{budzianowski2024pheme}
P.~Budzianowski, T.~Sereda, T.~Cichy, and I.~Vulic, ``Pheme: Efficient and conversational speech generation,'' 2024.

\bibitem{ju2024naturalspeech}
Z.~Ju, Y.~Wang, K.~Shen, X.~Tan, D.~Xin, D.~Yang, Y.~Liu, Y.~Leng, K.~Song, S.~Tang, Z.~Wu, T.~Qin, X.-Y. Li, W.~Ye, S.~Zhang, J.~Bian, L.~He, J.~Li, and S.~Zhao, ``Naturalspeech 3: Zero-shot speech synthesis with factorized codec and diffusion models,'' 2024.

\bibitem{li2023styletts}
Y.~A. Li, C.~Han, V.~S. Raghavan, G.~Mischler, and N.~Mesgarani, ``Styletts $2$: Towards human-level text-to-speech through style diffusion and adversarial training with large speech language models,'' 2023.

\bibitem{chen2021speech}
L.~Chen, Y.~Deng, X.~Wang, F.~K. Soong, and L.~He, ``Speech bert embedding for improving prosody in neural tts,'' 2021.

\bibitem{Liu_SSL_prosody}
Z.-C. Liu, Z.-H. Ling, Y.-J. Hu, J.~Pan, J.-W. Wang, and Y.-D. Wu, ``Speech synthesis with self-supervisedly learnt prosodic representations,'' pp.~7--11, 08 2023.

\bibitem{zhong2023eetts}
Y.~Zhong, C.~Zhang, X.~Liu, C.~Sun, W.~Deng, H.~Hu, and Z.~Sun, ``Ee-tts: Emphatic expressive tts with linguistic information,'' 2023.

\bibitem{guo2022prompttts}
Z.~Guo, Y.~Leng, Y.~Wu, S.~Zhao, and X.~Tan, ``Prompttts: Controllable text-to-speech with text descriptions,'' 2022.

\bibitem{guo2023prompttts2}
Y.~Leng, Z.~Guo, K.~Shen, X.~Tan, Z.~Ju, Y.~Liu, Y.~Liu, D.~Yang, L.~Zhang, K.~Song, L.~He, X.-Y. Li, S.~Zhao, T.~Qin, and J.~Bian, ``Prompttts 2: Describing and generating voices with text prompt,'' 09 2023.

\bibitem{Latif_e2e_prosody}
S.~Latif, I.~Kim, I.~Calapodescu, and L.~Besacier, ``Controlling prosody in $end-to-end$ tts: A case study on contrastive focus generation,'' pp.~544--551, 01 2021.

\bibitem{stephenson22_interspeech}
B.~Stephenson, L.~Besacier, L.~Girin, and T.~Hueber, ``{BERT, can HE predict contrastive focus? Predicting and controlling prominence in neural TTS using a language model},'' in {\em Proc. Interspeech 2022}, pp.~3383--3387, 2022.

\bibitem{intsint_1}
A.~Louw and E.~Barnard, ``Automatic intonation modeling with intsint,'' 11 2004.

\bibitem{intsint_tobi}
M.~M.~G. Leonardo Contreras~Roa, ``Un système d’équivalences entre tobi et intsint pour l’étude de l’interlangue prosodique,'' 12 2018.

\bibitem{TL_ToBI}
E.~Estebas, ``Tl tobi: A new system for teaching and learning intonation,'' 08 2013.

\bibitem{Hirst_momel}
D.~Hirst and R.~Espesser, ``Automatic modelling of fundamental frequency curves.,'' p.~1480, 01 1989.

\bibitem{hirst1998intonation}
D.~Hirst and A.~di~Cristo, {\em Intonation Systems: A Survey of Twenty Languages}, vol.~76.
\newblock 01 1998.

\bibitem{Roll_2023}
N.~Roll, C.~Graham, and S.~Todd, ``Psst! prosodic speech segmentation with transformers,'' in {\em Proceedings of the 27th Conference on Computational Natural Language Learning (CoNLL)}, Association for Computational Linguistics, 2023.

\bibitem{peng2024empirical}
Y.~Peng, I.~Kulikov, Y.~Yang, S.~Popuri, H.~Lu, C.~Wang, and H.~Gong, ``An empirical study of speech language models for prompt-conditioned speech synthesis,'' 2024.

\bibitem{Rosenberg2010AutoBI}
A.~Rosenberg, ``Autobi - a tool for automatic tobi annotation,'' in {\em Interspeech}, 2010.

\bibitem{bain2023whisperx}
M.~Bain, J.~Huh, T.~Han, and A.~Zisserman, ``Whisperx: Time-accurate speech transcription of long-form audio,'' 2023.

\bibitem{Wei2023}
H.~Wei, X.~Cao, T.~Dan, and Y.~Chen, ``Rmvpe: A robust model for vocal pitch estimation in polyphonic music,'' in {\em INTERSPEECH2023}, ISCA, aug 2023.

\bibitem{stress_nucleus_1}
C.~Gussenhoven, ``Stress shift and the nucleus,'' {\em Linguistics}, vol.~21, pp.~303--340, 01 1983.

\bibitem{Hart_Collier_Cohen}
J.~T. Hart, R.~Collier, and A.~Cohen, {\em A Perceptual Study of Intonation: An Experimental-Phonetic Approach to Speech Melody}.
\newblock Cambridge Studies in Speech Science and Communication, Cambridge University Press, 1990.

\bibitem{knyazev_evstigneeva}
S.~Knyazev and M.~Evstigneeva, ``“word-by-word” melodic contour in russian dialects: quantitative approach,'' pp.~284--294, 06 2022.

\bibitem{lobanov_1}
B.~Lobanov, ``Comparison of melodic portraits of english and russian dialogic phrases,'' {\em Computational Linguistics and Intellectual Technologies: Proceedings of the International Conference 'Dialogue 2016'}, vol.~517, no.~3, pp.~4054--4069, 2016.

\bibitem{ode2008transcription}
C.~ODÉ, ``Transcription of russian intonation tori, an interactive research tool and learning module on the internet,'' {\em Studies in Slavic and General Linguistics}, vol.~34, 01 2008.

\bibitem{o'connor1973}
J.~O'Connor and G.~Arnold, {\em The Intonation of Colloquial English}.
\newblock 01 1973.

\bibitem{c6906d38f66b4881848463e3eef3f3dc}
S.~Kakouros and J.~O'Mahony, ``What does bert learn about prosody?,'' in {\em Proceedings of the 20th International Congress of Phonetic Sciences} (R.~Skarnitzl and J.~Vol{\'i}n, eds.), pp.~1454--1458, Guarant International, 08 2023.

\bibitem{devlin2019BERT}
J.~Devlin, M.-W. Chang, K.~Lee, and K.~Toutanova, ``Bert: Pre-training of deep bidirectional transformers for language understanding,'' in {\em North American Chapter of the Association for Computational Linguistics}, 2019.

\bibitem{libri_tts}
H.~Zen, V.~Dang, R.~Clark, Y.~Zhang, R.~Weiss, Y.~Jia, Z.~Chen, and Y.~Wu, ``Libritts: A corpus derived from librispeech for text-to-speech,'' pp.~1526--1530, 09 2019.

\bibitem{Phoneme_bert}
Y.~A. Li, C.~Han, X.~Jiang, and N.~Mesgarani, ``Phoneme-level bert for enhanced prosody of text-to-speech with grapheme predictions,'' {\em ICASSP 2023-2023 IEEE International Conference on Acoustics, Speech and Signal Processing (ICASSP)}, pp.~1--5, 2023.

\end{thebibliography}

\end{document}